# AI-Driven MRI-based Brain Tumour Segmentation Benchmarking

Connor Ludwig
Engineering Science
University of Toronto
Toronto, Canada
connor.ludwig@mail.utoronto.ca

Khashayar Namdar
Institute of Medical Science
University of Toronto
Toronto, Canada
ernest.namdar@utoronto.ca

Farzad Khalvati
Department of Medical Imaging
University of Toronto
Toronto, Canada
farzad.khalvati@utoronto.ca

*Abstract*—Medical image segmentation has greatly aided medical diagnosis, with U-Net based architectures and nnU-Net providing state-of-the-art performance. There have been numerous general promptable models and medical variations introduced in recent years, but there is currently a lack of evaluation and comparison of these models across a variety of prompt qualities on a common medical dataset. This research uses Segment Anything Model (SAM), Segment Anything Model 2 (SAM 2), MedSAM, SAM-Med-3D, and nnU-Net to obtain zero-shot inference on the BraTS 2023 adult glioma and pediatrics dataset across multiple prompt qualities for both points and bounding boxes. Several of these models exhibit promising Dice scores, particularly SAM and SAM 2 achieving scores of up to 0.894 and 0.893, respectively when given extremely accurate bounding box prompts which exceeds nnU-Net's segmentation performance. However, nnU-Net remains the dominant medical image segmentation network due to the impracticality of providing highly accurate prompts to the models. The model and prompt evaluation, as well as the comparison, are extended through fine-tuning SAM, SAM 2, MedSAM, and SAM-Med-3D on the pediatrics dataset. The improvements in point prompt performance after fine-tuning are substantial and show promise for future investigation, but are unable to achieve better segmentation than bounding boxes or nnU-Net.

## I. Introduction

Medical image segmentation has revolutionized analysis of medical imaging by providing an accurate and efficient method of annotating regions of interest (ROI) [1]. Multiple benchmark datasets and competitions exist that aim to evaluate and compare deep learning (DL) models for medical image segmentation. The Brain Tumour Segmentation Challenge (BraTS) [2] serves as a widely recognized benchmark, offering a robust framework for the evaluation and comparison of diverse pipelines and DL architectures. In this research, both the BraTS 2023 adult glioma dataset and the BraTS 2023 pediatric dataset were used to train and validate segmentation models, investigating generalizability across both adult and pediatric populations and enabling robust evaluation on unseen data.

A wide range of architectures has been proposed for medical image segmentation, with U-Net [3] being the most commonly used in segmentation challenges. U-Net is a fully convolutional network (FCN) that features an encoder-decoder structure: the encoder functions as a conventional convolutional neural network (CNN) for hierarchical feature extraction, while the decoder employs up-convolutions to reconstruct spatial resolution and generate segmentation maps [4]. The success of U-Net in image segmentation has led to the development of several models that are extensions of the U-Net architecture, such as nnU-Net [5]. nnU-Net was proposed to improve the generalizability of image segmentation models. The model self-tunes its hyperparameters to act as an out-of-the-box image segmentation model that can be trained on any dataset. Another architecture of note is the Vision Transformer (ViT), which relies on transformers to enhance the ability to model long-range dependencies, something that traditional U-Net and other CNN-based architectures struggle with [6]. Segment Anything Model (SAM) is a recent architecture, proposed as an out-of-the-box 2D image segmentation that works on any segmentation task. SAM relies on the ViT architecture combined with the ability to be prompted with points or bounding boxes, providing a powerful and general model [7]. SAM's success as a general image segmentation model has inspired several models based on it. One of these models is MedSAM, proposed as a 2D image segmentation model that works on any medical imaging task [8]. Segment Anything Model 2 (SAM 2) is another segmentation model based on the SAM architecture, proposed as an out-of-the-box image model with performance improvements over SAM and the ability to work on video input [9]. Another model of note is SAM-Med3D, which adapts the SAM architecture to perform prompted volumetric image segmentation, working both out-of-the-box or fine-tuned on a task [10]. This research benchmarks these leading architectures (U-Net, nnU-Net, and SAM-based models including MedSAM, SAM 2, and SAM-Med3D) on brain tumour segmentation tasks to evaluate their performance and generalizability across both adult and pediatric datasets.

Although multiple benchmark models exist for medical image segmentation tasks, there is a gap in comprehensive model evaluation and comparison. Additionally, benchmark competitions such as BraTS withhold image masks at test time [10]. Thus, promptable models cannot be implemented for these competitions. This leaves an absence of comprehensive comparison between SAM-based models and traditional U-Net based architectures. This research aims to benchmark image segmentation models on the BraTS 2023 adult glioma and BraTS 2023 pediatric tumours datasets, using public training set mask data to generate model prompts, capturing zero-shot and fine-tuned performance for comparison among models [2]. Further, this research will use different prompting styles to evaluate the impact of prompt quality of performance in promptable segmentation models. This will provide an aggregate of state-of-the-art image segmentation performance from both promptable and non-promptable models on the same medical dataset for easy comparison. This collection of models performance will inform other researchers in model selection for medical segmentation tasks.

Key contributions of this research include:

- Evaluation and comparison of Dice scores of state-of-the-art image segmentation models on BraTS 2023 adult glioma and pediatrics datasets.
- Evaluation of the impact of prompt quality on promptable image segmentation models on medical images.
- Analysis of effectiveness of fine-tuning general image segmentation models to a medical segmentation task

## II. Related Works

Image segmentation is a thoroughly explored area of computer vision that has numerous applications [11]. It involves partitioning an image into meaningful regions or objects that collectively define its structure. Image segmentation has played a vital role in medical image analysis as 70% of challenges regarding medical image analysis rely on image segmentation [12]. DL-based medical image segmentation has proven to be vital in identifying and annotating ROIs across a variety of organs, achieving near human levels of accuracy [4]. In contrast to manual segmentation of medical images that are time-consuming processes, DL-based approaches to segmentation can reduce the labor required and improve consistency [13].

Several models have been introduced to achieve high quality image segmentation, particularly in biomedical images. One architecture that has seen exceptional success is U-Net [6]. U-Net is built upon CNNs and uses traditional convolutional and max pooling layers to extract spatial information from an image and then uses up-convolution to generate the segmentation mask [4]. The U-Net architecture is only designed for 2-dimensional input, but its success in image segmentation spawned other models based on its architecture including a 3D U-Net model, enabling accurate automated 3 dimensional image segmentation [14].

The U-Net architecture has been extended to create nnU-Net [4], designed to configure itself during training, avoiding manual hyperparameter selection in training, capable of adapting to any provided dataset. nnU-Net is versatile and can be trained on any 2D or 3D medical imaging dataset with any number of channels. nnU-Net has achieved state-of-the-art performance across a variety of medical datasets (33 out of the 53 segmentation tasks). nnU-Net has been extremely successful in brain tumour segmentation and was used as part of an ensemble in the winning solution in BraTS 2020, 2021, 2022, and 2023 [15], [16], [17], [18].

While U-Net based architectures have been dominant in medical image segmentation, a recently introduced model, SAM, has demonstrated excellent generalized performance on non-medical datasets [6]. SAM is a promptable 2D segmentation network. Thus, it can either be used to generate masks for the full image or it can be prompted with a bounding box and/or points. Despite the superior generalized performance of SAM on natural images, its performance ranges wildly on medical tasks and can result in poor performance [19].

Additionally, the performance of SAM significantly depends on prompt type (e.g., precise versus rough ROI bounding box). The 2D architecture of SAM is another limitation which is crucial on medical datasets, as 3D images must be passed one slice at a time. Further, SAM is limited to 3 input channels, meaning that in a dataset such as BraTS that contains 4 magnetic resonance imaging (MRI) sequences per patient, only 3 can be used as input.

Building upon the SAM architecture, Meta introduced SAM 2 that is capable of video segmentation and unifies image and video segmentation by treating images as a single frame from a video [9]. To segment a video, a prompt for any number of frames can be provided, and then the model will propagate masks through all frames. In addition to introducing video segmentation, SAM 2 shows performance improvements in both intersection over union (IOU) score and segmentation speed, when compared to SAM's image segmentation. When SAM 2 image segmentation is applied to 3D CT and MRI datasets, it performs better than SAM, but worse than nnU-Net, although the relation between prompt and performance is not properly explored [20].

The impressive general performance of SAM inspired the creation of a general out-of-the-box promptable model for medical image segmentation, leading to MedSAM [8]. This model was designed to accept 2D images with up to 3 channels, rendering it unable to capture relationships between slices as opposed to a 3D model. Additionally, there are limitations on performance when making a fully general medical image segmentation model due to the great diversity between medical datasets combined with a variety of complex relationships within images [5]. Despite its limitations, MedSAM achieves high Dice scores on unseen segmentation tasks and can outperform specialized models on some datasets.

Addressing the limitations of 2D architectures, SAM-Med3D was introduced. Based on the SAM architecture, it is able to achieve accurate segmentation with as few as one prompt for an entire volume in contrast to 2D approaches requiring one per slice. Additionally, the 3D approach allows capturing inter-slice correlations, which increases the information available to the model and improves inter-slice consistency when compared to 2D approaches.

When comparing segmentation models for medical images, international challenges have become the standard for comparison, providing a common dataset for fair model comparison [12]. However, the format of medical image segmentation competitions restricts the ability of promptable models to compete. At test time, the masks are withheld, preventing prompts from being automatically generated.

There have been several evaluations of promptable segmentation models performance [7], [8], [9], [10], but these evaluations often occur on different datasets, use different prompting styles, and/or use different evaluation metrics. These combinations of factors make it difficult to directly compare such models. In contrast, this research evaluates all previously mentioned models using a variety of prompting styles across the same dataset to conduct a fair and comprehensive comparison.

## III. METHODOLOGY

### A. Datasets

For this research, both the BraTS 2023 adult glioma dataset and the BraTS 2023 pediatrics dataset were used. The BraTS 2023 adult glioma dataset was selected due to its nature as a large, commonly used, dataset composed of MRI scans of adult brain tumours [2]. Its training set is identical to the BraTS 2021 training set and consists of 1251 patients each with four MRI sequences, including native (T1-weighted), post-contrast T1-weighted (T1Gd), T2-weighted (T2), and T2 Fluid Attenuated Inversion Recovery (FLAIR), and additionally multiple ROI ground truth segmentation masks are available for each patient. The ground truth mask is only provide to the public for the training set, meaning that only the training set can be used for promptable model evaluation and comparison. BraTS makes use of a thorough pre-processing pipeline in which the brain is aligned to a standard template (i.e, SRI24 atlas [21]), the scan is resampled to a constant size per voxel and the skull is stripped from the scans. Despite the factors that make BraTS an ideal dataset for model evaluation and comparison, there is a concern regarding data leakage. Data leakage occurs when a model is tested on data included in the training set. This inflates performance, making comparison unfair [21]. Some general-purpose medical models, such as MedSAM included BraTS in their training set, meaning that data leakage may occur when comparing model performance on the BraTS dataset. The BraTS 2023 Pediatrics dataset was selected for evaluation in addition to the BraTS 2023 adult glioma dataset to resolve the data leakage issue as the pediatrics dataset was not used to train any models being evaluated. The BraTS 2023 pediatrics dataset contains the same MRI sequences, resolution, and pre-processing pipeline as the main BraTS 2023 dataset enabling any trained on the BraTS 2023 adult dataset to be directly applied on the pediatrics dataset [23]. The pediatrics dataset is significantly smaller than the BraTS 2023 adult dataset. It only contains 99 patients in the training set with a multiple ROI segmentation masks available. This reduced dataset size introduces a difficulty in training any model on the pediatrics dataset that is not present when training on the adult dataset [24].

### B. Data Processing

The models being benchmarked span a wide range of architectures, each with their own constraints on input to the model requiring different data processing. The variance in input constraints caused difficulty in ensuring fair comparison, but several steps were taken to ensure proper benchmarking.

In this work, five models will be benchmarked on brain tumour segmentation tasks to evaluate their performance and generalizability across both adult and pediatric datasets. These models include: U-Net, nnU-Net, and SAM-based models including MedSAM, SAM 2, and SAM-Med3D. Of the 5 models being benchmarked, the input constraints can be subdivided into 3 categories: 2D vision transformers, including SAM, SAM 2, and MedSAM; 3D vision transformers, including SAM-Med-3D; and 3D U-Net, including nnU-net.

The 3D vision transformer architecture employed by SAM-Med-3D is well suited to this task with some limitations. The model can only accept 3D images with one channel, meaning that it is unable to make use of all 4 MRI sequences provided in the BraTS 2023 Adult and Pediatrics dataset [10]. Due to this limitation, the FLAIR sequence was passed as input to the model, due to past literature indicating the flair sequence yielding the best segmentation results in cases of single channel segmentation [25]. One further limitation on this model is that it has only been designed to accept point prompts for image segmentation. Thus, this research benchmarked this model with varying quality of point prompts, but data for box prompts could not be collected.

The 2D vision transformer architecture used by SAM, SAM 2, and MedSAM, comes with its own set of input limitations. Each of these models is designed to only accept 1024 by 1024 pixel 2D images with 3 channels [7], [8], [9]. To circumvent this input limitation, the MRI sequences for each patient were sliced along the axial view, yielding a sequence of 155 2D images, each 240 by 240 pixels. Although the models are capable of receiving 3 channels of information in their input, for fairness of comparison against SAM-Med-3D, only 1 sequence was used in model evaluation. As with SAM-Med-3D, the FLAIR sequence was selected due to literature indicating its status as the best singular sequence to segment [25]. To facilitate the use of a single sequence image with 240 by 240 pixels for these models, the FLAIR sequence was replicated to form 3 identical channels, then each slice was upscaled to 1024 by 1024. Both SAM and SAM 2 were designed to accept point and box prompts, allowing data collection with both prompting style. MedSAM was initially designed for solely box prompts and was later updated to accept point prompts, although the prompt encoder could not be configured to handle more than one point. Thus, only box prompt data and single point prompt data were able to be collected for MedSAM.

The 3D U-Net architecture employed by nnU-Net can be trained on any number of channels for a 3D image of any size. However, in this research, a pre-trained model trained on BraTS 2021 was used to perform inference on the MRI scans [16]. Due to the pre-trained nature of this model, it can only accept 3D images that are 240 by 240 by 155 voxels with 4 channels. To comply with the limitations on the model input, the model was provided all sequences with no image upscaling required. This approach was chosen rather than replicating the FLAIR sequence 4 times because this model was specifically trained on the pixel intensities found in each sequence. This method results in more sequences being passed to nnU-Net than the models its being compared against. Due to the non-promptable nature of nnU-Net, no prompting data was able to be collected and instead one un-prompted trial was run across both datasets.

### C. Prompt styles

A set of prompting methods were established for each prompt type to achieve fair comparison across prompt styles and quality could be established. For each prompting style, points and boxes, 3 varying quality of prompt were tested, a high-quality prompt, a medium-quality prompt, and a low-quality prompt. The high-quality prompt was intended to represent a best-case scenario, while the medium and low-quality prompts were intended to emulate the prompts a human might provide in a clinical setting. The box prompts used were sampled from ground truth masks with varying size scaling and positional shift applied to the box. For the high-quality prompt, the bounding box was the smallest possible box that fully captured the tumour. For the medium-quality prompt, each dimension of the box was increased in size by 10%, then it was shifted randomly up or down by up to 10% of its height sampled from a uniform distribution and shifted to either side by up to 10% of its width sampled from a uniform distribution. This same scaling and shift process was applied to the low-quality prompt, except it was scaled by 30% and shifted by up to 20% of its size. Sample bounding box prompts overlayed with ground truth segmentation masks can be seen in Fig. 1. The point prompts used were randomly sampled from the ground truth segmentation mask of the patient, with a varying number of points provided depending on prompt quality. When providing point prompts to the promptable models, more points result in more information, thus the minimal information is from 1 point and it was chosen as the low-quality prompt. 10 points was chosen as the high-quality prompt due to use in literature [10], and 5 was chosen as the medium-quality point prompt to act as a midway between the high-quality and low-quality prompt.

For each of SAM, SAM 2, and MedSAM, the prompts had to be passed on a per slice basis due to the 2D nature of the models. This means that to complete the segmentation of a patient with box prompts, N bounding boxes must be drawn where N is the number of slices with tumour region in the patient. However, SAM 2's video segmentation mode can be used to propagate masks through a series of images, enabling segmentation to be performed without masks on each slice [9]. When using the video segmentation mode, a prompt was provided on the first slice with any tumour region, and then every 10th slice with tumour region after that. This prompting method results in providing prompts on [N/10] slices rather than N. SAM-Med-3D requires prompts per volume rather than per slice due to its 3D nature. This results in only providing either 1-, 5-, or 10-point prompts per patient depending on if the low-, medium-, or high-quality point prompting method is being used. The number of point prompts provided per patient is independent of how many slices contain tumour regions and is always either 1, 5, or 10.

### D. Evaluation Metrics

To evaluate quality of segmentation, the Dice Similarity Coefficient (DSC or Dice score for short) was employed due to its frequent use in literature and medical image segmentation competitions [3]. To perform accurate DSC calculations and fair comparison, all DSC scores were calculated on the 3D scan rather than 2D slices. To achieve this for 2D based segmentation models, the predicted masks for each slice were saved and then compiled into one 3D segmentation mask. Inference time was measured to determine the efficiency of the model at patient segmentation. Additionally, training time was tracked to evaluate the computational resources required to fine-tune these models for this task.

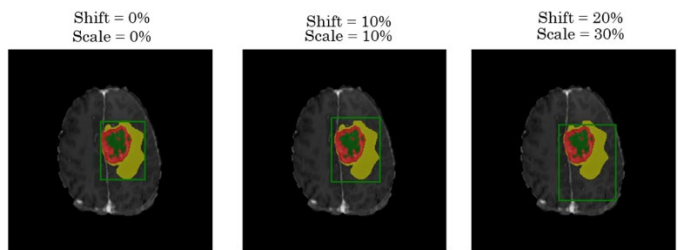

Fig. 1. High-, mid-, and low-quality bounding box prompts overlayed with the ground truth mask.

### E. Model Fine-Tuning

All SAM based models were fine-tuned on the BraTS 2023 pediatrics dataset to determine how well they could be adapted to this medical segmentation task. A modified 4-fold cross validation approach was used with a 50% training, 25% validation and 25% test split as shown in Fig. 2. This training, validation, and test split allowed a hyperparameter search to be conducted without the computational burden of conducting a full nested cross validation training procedure. Each model had checkpoints saved at the end of each epoch, the models were fine-tuned with early stopping when the validation Dice score stopped decreasing and the checkpoint with highest validation score was saved. The final hyperparameter combination for each model was chosen based on the average of its best validation scores across each fold. Then for the selected hyperparameter combination, its best model checkpoint for each fold was loaded and used to obtain inference across all applicable prompt styles on its respective test fold. Additionally, the fine-tuned SAM 2 model was used to obtain SAM 2 video test results.

To fine-tune SAM, SAM 2, and MedSAM, all patient scans were preprocessed and slices saved as Joint Photographic Experts Group (jpg) files prior to beginning training. All jpg files belonging to the same patient were kept in the same fold, and only slices containing tumour regions were used for the training and evaluation process. Each slice was randomly provided either a single point prompt or a perfect bounding box prompt and then used to train the model. For these models, Validation DSC was computed by running the model across all slices in the validation fold and using either a single point prompt or a perfect bounding box to prompt the model with a 50% chance of either. The resulting validation Dice score was the average Dice score of each slice. This computation does not perfectly mirror the proper 3D Dice score calculation, but it provides similar results. The fine-tuning process for SAM-Med-3D was extremely similar to that of SAM, SAM 2, and MedSAM, with a couple differences to compensate for the 3D architecture. Rather than saving the images as jpg, they are kept in their original file format. Because SAM-Med-3D is only promptable with points, for both training and validation, the prompt was either 1, 5, or 10 points with equal probability. The Dice score for each patient in the validation fold was computed with the 3D generated masks, rather than using the average of 2D slices like in the validation score for the other models. All codes are available at: https://github.com/IMICSLab/Brain_Tumor_Segmentation.

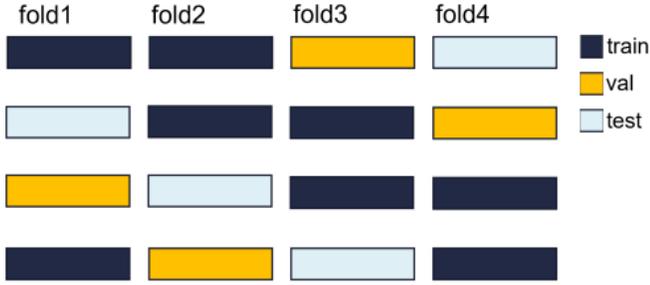

Fig. 2. 4-fold cross validation scheme used for model fine-tuning.

## IV. RESULTS

### A. BraTS 2023 Adult Gioma Zero Shot Inference

All models were used to perform zero-shot segmentation across the entire BraTS 2023 Adult Glioma training set according to the method laid out previously in this research, capturing all prompt data for promptable models except box prompts for SAM-Med-3D and 5 and 10 points for MedSAM due to limitations with prompt encoder design. Table I shows the average Dice scores obtained by all models across all applicable prompting methods. Of these results, MedSAM, SAM-Med-3D and nnU-net all benefit from data leakage to varying degrees. nnU-Net has the highest Dice score across all models and prompting styles with a score of 0.958 despite being unprompted. SAM 2 demonstrates the highest Dice score across each level of box prompt quality, with both the high and medium quality boxes achieving high Dice scores approaching 0.9. Across all prompt styles, SAM 2's video segmentation mode obtains slightly worse Dice scores than SAM 2's image segmentation mode. SAM is outperformed by SAM 2 on all box prompt qualities, but outperforms SAM 2 on point prompts. SAM-Med-3D achieved the highest Dice score at each level of point prompt quality, obtaining a score of 0.815 with the highest quality prompt and 0.730 with the lowest quality. MedSAM exhibits the worst Dice score results of all models, obtaining a score of 0.685 with a high quality box prompt.

### B. BraTS 2023 Pediatrics Zero Shot Inference

Zero-shot segmentation was performed across the entire BraTS 2023 Pediatrics training set with every model according to the method laid out previously in this research. Table II shows the average Dice scores obtained by all models across all applicable prompting methods. Unlike with the adult glioma dataset, there is no data leakage for these results. With no data leakage, the Dice score of nnU-Net drops to 0.860. On this dataset SAM 2 prompted with a high-quality box exhibits the best performance across all methods regardless of prompting method, achieving a Dice score of 0.886. SAM 2 exhibits the best results at both high-quality and medium-quality box prompts but is surpassed by SAM when provided a low-quality box prompt. Similar to the adult glioma dataset, SAM 2's image segmentation mode slightly outperforms its video segmentation mode. Additionally, SAM outperforms SAM 2 on all point prompt qualities, like on the adult glioma dataset. SAM-Med-3D has the highest Dice score at each level of point prompt quality. MedSAM obtained the lowest Dice Score of all models, reaching its highest Dice score of 0.633 with a medium quality box prompt.

TABLE I. BRATS 2023 ADULT GLIOMA ZERO-SHOT DSC

|  | 1 Point | 5 Points | 10 Points | High Quality Box | Medium Quality Box | Low Quality Box |
|---|---|---|---|---|---|---|
| SAM | 0.325 | 0.131 | -0.015 | 0.017 | 0.007 | -0.019 |
| SAM 2 | 0.363 | 0.277 | 0.152 | 0.007 | 0.019 | 0.060 |
| SAM 2 Video | 0.280 | 0.128 | 0.007 | 0.015 | 0.029 | 0.064 |
| MedSAM | 0.194 | N/A | N/A | 0.054 | 0.030 | 0.007 |
| SAM-Med-3D | 0.014 | 0.007 | 0.000 | N/A | N/A | N/A |

| Model | Box Prompt Quality ||| Point Prompts ||| Unprompted |
|---|---|---|---|---|---|---|---|
|  | High | Medium | Low | 10 points | 5 points | 1 point |  |
| SAM | 0.869 | 0.783 | 0.688 | 0.605 | 0.453 | 0.295 | N/A |
| SAM 2 | 0.893 | 0.876 | 0.759 | 0.474 | 0.393 | 0.285 | N/A |
| SAM 2 Video | 0.835 | 0.792 | 0.685 | 0.459 | 0.330 | 0.166 | N/A |
| MedSAM | 0.685 | 0.654 | 0.606 | N/A | N/A | 0.050 | N/A |
| SAM-Med-3D | N/A | N/A | N/A | 0.815 | 0.793 | 0.730 | N/A |
| nnU-Net | N/A | N/A | N/A | N/A | N/A | N/A | **0.958** |

TABLE II. BRATS 2023 PEDIATRICS ZERO-SHOT DSC

| Models | Box Prompt Quality ||| Point Prompts ||| Unprompted |
|---|---|---|---|---|---|---|---|
|  | High | Medium | Low | 10 points | 5 points | 1 point |  |
| SAM | 0.877 | 0.859 | 0.797 | 0.749 | 0.586 | 0.345 | N/A |
| SAM 2 | **0.886** | 0.862 | 0.777 | 0.623 | 0.474 | 0.322 | N/A |
| SAM 2 Video | 0.767 | 0.743 | 0.674 | 0.562 | 0.416 | 0.231 | N/A |
| MedSAM | 0.572 | 0.633 | 0.618 | N/A | N/A | 0.115 | N/A |
| SAM-Med-3D | N/A | N/A | N/A | 0.814 | 0.795 | 0.723 | N/A |
| nnU-Net | N/A | N/A | N/A | N/A | N/A | N/A | 0.860 |

### C. BraTS 2023 Pediatrics Fine-Tuning

All promptable models were fine-tuned on the BraTS 2023 pediatrics dataset using the training scheme detailed in the methods section, except for nnU-Net due to computational resource limitations. Once the models were trained, the best checkpoint of each model for each fold was used to obtain inference on its respective test fold, and their average Dice across all test folds can be seen in Table III. The relative performances of models after fine-tuning compared to before can be seen showcased in Fig. 3. After fine-tuning, SAM prompted with a high-quality bounding box obtained the highest Dice score, but SAM 2 maintained the best Dice score across the other 2 qualities of box prompt. When prompted with a high or medium quality box prompt, both SAM and SAM 2 surpassed the Dice score of nnU-Net. Similar to both zero shot trials, SAM 2's video segmentation mode was outperformed by its image segmentation mode and MedSAM performed the worst across all prompting methods. SAM-Med-3D demonstrated the best performance across all qualities of point prompt, achieving a Dice score of 0.814 with a high quality point prompt.

*D. Computational Intensity*

Zero-shot inference on the BraTS 2023 Adult Glioma dataset was timed to obtain whole dataset inference and per-patient inference times. All timing data was collected when using the inference scripts found in the model's respective repositories on a single L4 GPU. Inference times across prompt styles were found to be roughly constant. Thus, the results were summarized into one number for each model and can be seen in Table IV. MedSAM saw the slowest inference time due to its costly data preprocessing that had to be performed on every slice, taking 2 minutes per patient. SAM-Med-3D saw the fastest inference time due to being a lighter model and segmenting the entire volume at once, segmenting scans in just 0.80 seconds per patient. SAM 2 had the second fastest inference time at 7.2 seconds per patient, but this increased greatly to 19 seconds per patient when used in video segmentation mode. Both SAM 2 image and video segmentation ran faster than nnU-Net which took 22 seconds per patient. SAM was slower than all models except MedSAM, taking 72 seconds per patient.

Fig. 3. Relative improvements of fine-tuned Dice scores relative to zero-shot Dice scores

TABLE III. BRATS 2023 PEDIATRICS FINE-TUNED DSC

| Models | DSC | | | | | |
|---|---|---|---|---|---|---|
| | Box Prompt Quality | | | Point Prompts | | |
| | High | Medium | Low | 10 points | 5 points | 1 point |
| SAM | **0.894** | 0.866 | 0.777 | 0.734 | 0.717 | 0.670 |
| SAM 2 | 0.893 | 0.881 | 0.837 | 0.775 | 0.751 | 0.686 |
| SAM 2 Video | 0.782 | 0.772 | 0.737 | 0.568 | 0.543 | 0.511 |
| MedSAM | 0.626 | 0.664 | 0.626 | N/A | N/A | 0.309 |
| SAM-Med-3D | N/A | N/A | N/A | 0.814 | 0.802 | 0.737 |

During training, each epoch and total training time for the model was recorded when run on a single A100 GPU. Table V shows the average epoch and model training times per model. The epoch times held roughly constant within the same model, but the model training times varied much more due to variance in the number of epochs until early stopping occurred. SAM-Med-3D had the fastest training time due to its lightweight architecture and segmentation of whole volumes in one pass taking only 1.5 minutes per epoch. SAM 2 and MedSAM saw comparable training speeds, taking 7.5 and 10 minutes per epoch respectively. Due to its larger architecture, SAM had the slowest training speed, taking 25 minutes per epoch.

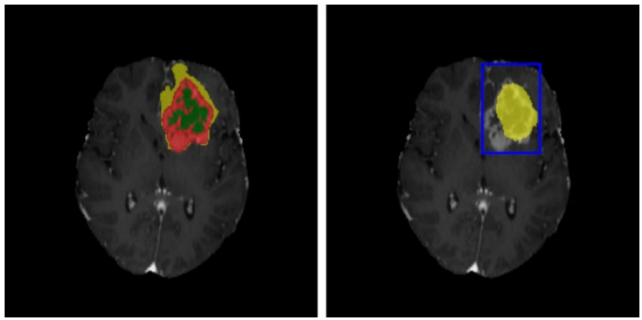

## V. Discussion

Results from the BraTS 2023 Pediatrics dataset indicate that fine-tuned versions of SAM and SAM 2 are capable of surpassing Dice scores from nnU-Net when given high or medium quality box prompts. However, both SAM and SAM 2 require many accurate prompts while nnU-Net operates automatically, hindering their ability to be used in a clinical setting. Both SAM 2 video segmentation and SAM-Med-3D allow fewer prompts than other SAM based models, but they are unable to outperform nnU-Net across all prompting methods tested. MedSAM exhibits poor results which runs contrary to what is expected from literature [8], [26] When visually observing the segmentation results of MedSAM in Fig. 4, one can notice that the predicted mask strays from the bounding box prompt and often resembles a circular blob not corresponding to any shapes in the scan itself. It is likely that in the original training of MedSAM it lost some of its generality and is attempting to find organ shapes in an image that does not have any.

Fig. 4. Left: Ground truth mask of brain tumour. Right: Predicted mask from MedSAM with accompanying bounding box prompt.

TABLE IV. MODEL INFERENCE TIME ON BRATS 2023 ADULT GLIOMA WITH AN L4 GPU

| Models | Inference Time | |
|---|---|---|
| | Whole Dataset (hours) | Per Patient (seconds) |
| SAM | 25 | 72 |
| SAM 2 | 2.5 | 7.2 |
| SAM 2 Video | 6.7 | 19 |
| MedSAM | 42 | 120 |
| SAM-Med-3D | **0.28** | 0.80 |
| nnU-Net | 7.8 | 22 |

TABLE V. MODEL TRAINING TIME ON BRATS 2023 PEDIATRICS WITH AN A100 GPU

| Models | Training Time | |
|---|---|---|
| | Model (hours) | Epoch (minutes) |
| SAM | 16 | 25 |
| SAM 2 | 4.7 | 7.5 |
| MedSAM | 5.3 | 10 |
| SAM-Med-3D | **0.65** | 1.5 |

When evaluating the impact of prompts on model performance, bounding box prompts emerge as the superior method across both datasets. These results align with expectations from literature, because point prompts possess more ambiguity than bounding boxes [8], [26]. All models were highly sensitive to the quality of prompt provided, even within the same prompt type. In zero shot inference, Dice scores with high quality box prompts were approximately 0.1 greater than low quality box prompts on average, and this difference was even greater in point prompts. When fine-tuning the models, the segmentation results for each prompt type improved, but the performance of point prompts improved substantially more than box prompts. There are likely two reasons that point prompts benefited the most from fine-tuning, they had a worse starting segmentation performance, and they struggled with ambiguity. Generally, lower quality prompts showed larger performance boost than higher quality prompts, which was true even within the same prompt type. For all models, single point segmentation improved more than 5 point which again improved more than 10 point. Additionally, SAM 2 and SAM 2 video results showed low-quality boxes seeing a bigger performance jump than high-quality boxes. This indicates that some of the enhanced performance increase in points after fine-tuning relative to box prompts can be simply attributed to their weaker starting point. However, visually inspecting the fine-tuned predicted masks, it can be observed that fine-tuning the model helped it resolve ambiguity. When a point was placed in a region such as the tumour core, the model would often segment just the core rather than the whole tumour, but as can be observed in Fig. 5, after fine-tuning, the model is able to better understand the objective is to segment the whole tumour and the predicted mask extends past just the tumour core.

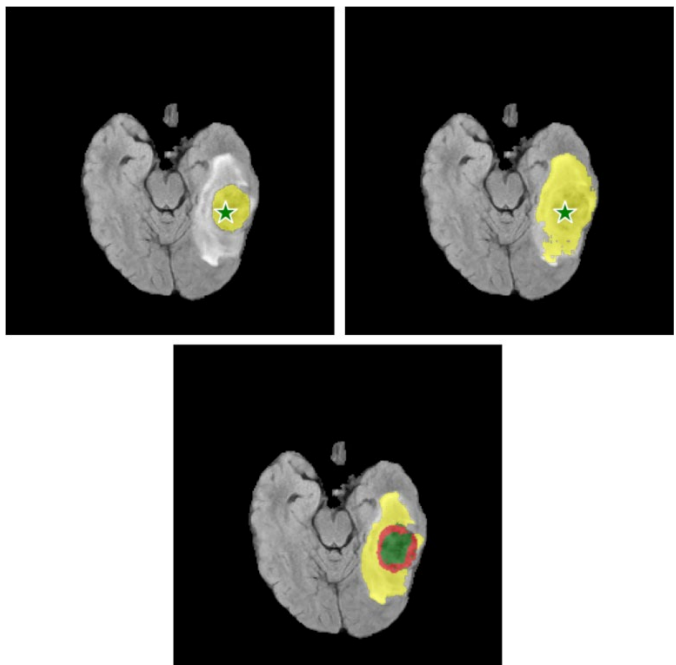

Fig. 5. SAM 2 pre-trained vs fine-tuned segmentation of brain tumour slice with single point prompt. Top Left is pre-trained, top right is fine-tuned, and bottom middle is the ground truth.